\documentclass[conference]{IEEEtran}

\IEEEoverridecommandlockouts
\usepackage{cite}
\usepackage{amsmath,amssymb,amsfonts}
\usepackage{algorithmic}
\usepackage{graphicx}
\usepackage{textcomp}
\usepackage{xcolor}
\usepackage{svg}
\usepackage{subcaption}

\def\BibTeX{{\rm B\kern-.05em{\sc i\kern-.025em b}\kern-.08em
    T\kern-.1667em\lower.7ex\hbox{E}\kern-.125emX}}
\usepackage{tikz}
\usepackage{url}
\usepackage{hyperref} 
\hypersetup{hidelinks,
    colorlinks=true,
    allcolors=black,
    pdfstartview=Fit,
    breaklinks=true
}
\usepackage{siunitx}
\usepackage{threeparttable}
\usepackage{multirow}
\usepackage{makecell}
\usepackage{booktabs}
\usepackage{float}
\usepackage{gensymb}
\usepackage[linesnumbered,ruled,vlined]{algorithm2e}
\usepackage{graphicx}
\usepackage{subcaption}
\usepackage{tabularx}
\usepackage[table]{xcolor}
\usepackage{colortbl}
\usepackage{array}
\usepackage{graphicx} % \resizebox
\usepackage{booktabs} % \toprule, \midrule, \bottomrule

\definecolor{darkgreen}{RGB}{0,100,0}

\SetCommentSty{emph}

\begin{document}
\title{
    ReWeight: Leveraging Human Data for 
    VLA \\Post-Training via Demonstration Retrieval and Sample Weighting
}

\author{
    Chenwei Wang,
    Dianye Huang, 
    Match W.L. Ko,
    Chenjia Bai,
   and Zhongliang Jiang 
    \thanks{
		% Manuscript received Month xx, 2xxx; revised Month xx, xxxx; accepted Month x, xxxx.
		% This work was supported in part by the xxx Department of xxx under Grant  (sponsor and financial support acknowledgment goes here).
        (\textit{Corresponding author: Zhongliang Jiang})
        
        Chenwei Wang, Dianye Huang, Match W.L. Ko, and Zhongliang Jiang are with the Department of mechanical Engineering, The University of Hong Kong, Hong Kong SAR, China. (e-mail: chenweiwang@connect.hku.hk, dianye.huang@hku.hk, matchko@hku.hk, zljiang@hku.hk). 
        
        Chenjia Bai is with the Institute of Artificial Intelligence (TeleAI), China Telecom, Shanghai 200232, China, and also with the Shenzhen Research Institute of Northwestern Polytechnical University, Xi’an 710072, China (e-mail: baicj@chinatelecom.cn).
    }
}

\maketitle
\begin{abstract}
Post-training vision-language-action (VLA) models for specific robots and tasks requires in-domain demonstrations, yet collecting diverse robot data is costly. Egocentric human demonstrations provide a scalable alternative, but directly mixing human and robot data can introduce cross-embodiment discrepancies and degrade policy performance. To address this challenge, we introduce ReWeight, a framework that incorporates human data into VLA post-training through demonstration-level retrieval and sample-level weighting. ReWeight learns a cross-embodiment visuomotor representation that combines visual observations with future actions to measure behavioral similarity between human and robot demonstrations. Based on optimal transport, it retrieves human demonstrations relevant to the target robot data and assigns larger weights to samples with smaller cross-embodiment discrepancies. We evaluate ReWeight using $\pi_{0.5}$ across eight simulation tasks and four real-world tasks under both clean and randomized settings. In simulation, ReWeight improves the average success rate of post-trained $\pi_{0.5}$ from 39\% with only robot data and 44\% with randomly mixed human-robot data to 57\%. In the physical experimental setting, it achieves an average success rate of 68.8\%, outperforming the baselines by 28.8\% and 13.8\%, respectively. 
% Under lighting changes and visual distractors, ReWeight achieves a 60.0\% overall success rate, compared with 28.8\% and 41.3\% for the two baselines. 
Overall, ReWeight provides an effective paradigm for transforming abundant egocentric human experience into transferable supervision for robot learning. (\textit{Project webpage:} \href{https://reweight-vla.github.io/}{https://reweight-vla.github.io})

% \footnote{Anonymous project webpage: \url{https://reweight-vla.github.io/}}

\end{abstract}


%%%%%%%%%%%%%%%%%% main content
% https://www.tablesgenerator.com/#  % create table
\newcommand{\fix}{\marginpar{FIX}}
\newcommand{\new}{\marginpar{NEW}}
\newcommand{\scr}[1]{{\scriptsize #1}}
\newcommand{\emphTab}[2]{{#1}\scr{(#2)}}
\newcommand{\up}[1]{\textcolor{upColor}{#1}}
\newcommand{\down}[1]{\textcolor{downColor}{#1}}
\newcommand{\revised}[1]{\textcolor{red}{#1}}

\section{Introduction}
\label{sec:intro}
\par
\IEEEPARstart{V}{ision}-language-action (VLA) models have shown strong promise for general-purpose robotic manipulation by integrating visual perception, language understanding, and action generation within a unified policy~\cite{o2024open,kim2024openvla}, compared to conventional control approaches~\cite{zhang2025novel,si2025deep,huang2026convex}. Scaling VLAs with diverse robot data has enabled increasingly broader generalization~\cite{o2024open,pmlr-v305-black25a}. However, deploying a pretrained VLA on a specific task or robot platform often relies on post-training with in-domain data to ground the policy in the target task, environment, and embodiment~\cite{kim2024openvla,pmlr-v305-black25a}. Therefore, the performance of a VLA remains closely tied to the availability and diversity of embodied data for adaptation.

\par
Collecting such data directly on robots requires task-specific hardware setup and repeated interaction with the environment, making it costly and difficult to acquire data with the scale and diversity demanded by the large models. In contrast, humans, representing the most prevalent embodiment in the physical world, can continuously generate diverse manipulation data at a scale far beyond what is feasible through robot data collection~\cite{hoque2026egodex,li2025scalable}. 
Human data has therefore attracted growing interest as a scalable and cost-effective source of embodied supervision for robot learning. In particular, EgoVLA and Being-H0 demonstrated that large-scale human videos can provide useful priors for subsequent adaptation to robotic platforms~\cite{yang2025egovla,luo2025being}. Besides, EgoMimic showed that jointly training on aligned human and robot demonstrations can substantially improve imitation learning~\cite{kareer2025egomimic}. More recently, Kareer~\emph{et al.} systematically showed that sufficiently large and diverse VLA pretraining can yield increasingly embodiment-agnostic representations, enabling human-to-robot transfer even without explicit cross-embodiment correspondence engineering~\cite{kareer2025emergence}. Being-H0.5 further extended it toward large-scale cross-embodiment learning across human and robotic embodiments~\cite{luo2026being}.

\begin{figure}[t]
    \centering
    \includegraphics[width=\columnwidth]{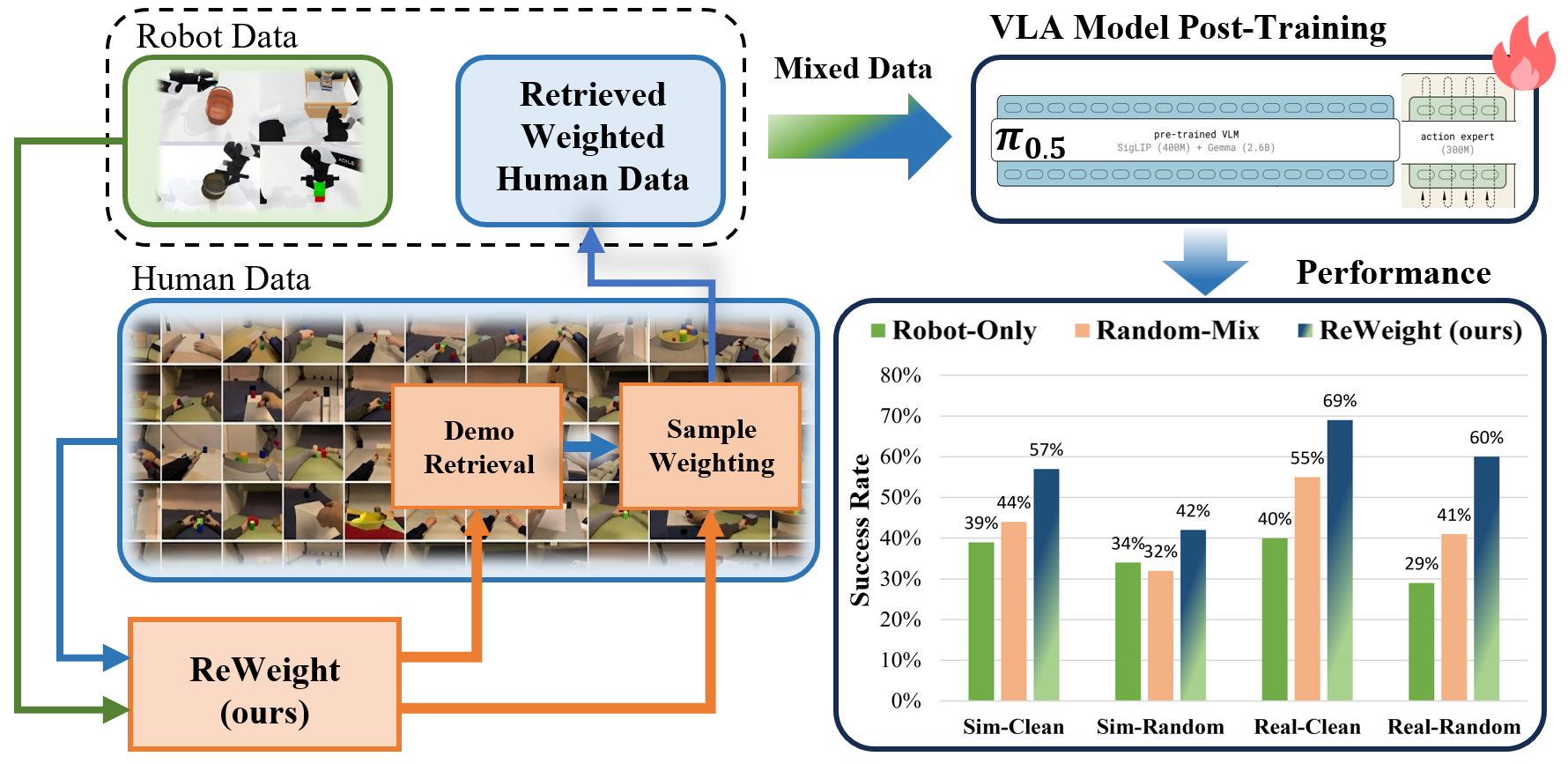}
    \caption{Illustration of ReWeight, retrieving human demonstrations and assigning sample weights for VLA post-training.
    % Illustraion of the proposed ReWeight, a framework that retrieves transferable human data and assigns sample-level weights for VLA post-training.
    }
    \label{fig:teaser}
\end{figure}

\par
However, simply scaling heterogeneous data does not guarantee better robot policies. Wang~\emph{et al.} showed that indiscriminately mixing data across embodiments can introduce interference and even degrade VLA performance~\cite{wang2026rethinking}. Similarly, Punamiya~\emph{et al.} found that the benefit of scaling human data depends critically on its alignment with the robot learning objective~\cite{punamiya2026egoverse}. These findings highlight the limitations of indiscriminate data scaling and motivate selective use of human data. Retrieval provides a direct mechanism for such selection. Prior approaches have exploited task similarity to retrieve relevant robot data from large offline datasets, including the use of lightweight human hand demonstrations as queries for task-relevant robot play data~\cite{hong2025hand}. Retrieval from large egocentric video banks has also been explored to provide task-relevant human data for robotic manipulation~\cite{zhu2025let}. SiMDex formulates human-data selection for VLA post-training as a recommendation problem and mines task-relevant egocentric samples from a large human-data pool, demonstrating advantages over randomly mixing human data~\cite{lin2026simdex}. 

\par
Yet retrieval alone cannot fully capture the varying relevance of human demonstrations to a target robot task. Even within the retrieved subset, demonstrations can differ substantially in their transferability to the target robot. Treating all retrieved samples equally overlooks this fine-grained variation and allows marginally relevant demonstrations to influence optimization as strongly as highly aligned ones. Sample weighting has been explored in robot learning to handle heterogeneous supervision, for example by emphasizing informative interventions or reweighting logged robot transitions during offline post-training~\cite{xie2025data,zhang2026conservative}. These methods, however, primarily estimate sample importance within the robot domain based on factors such as data quality or informativeness. In human-to-robot transfer, sample importance additionally depends on the cross-embodiment relevance of each human demonstration to the target robot behavior. Effective use of large-scale human data therefore requires jointly addressing two coupled decisions:  \textit{(i). which human demonstrations should be transferred?}, and \textit{(ii). how much should each selected sample contribute to VLA post-training?}

\par
To this end, we introduce ReWeight, a unified framework for leveraging human data for VLA post-training through demonstration-level retrieval and sample-level weighting. ReWeight first learns a cross-embodiment visuomotor representation that enables human and robot demonstrations to be compared despite differences in appearance and embodiment. Given a small set of robot data and a large pool of human data, ReWeight retrieves an equal size of human data that is most relevant to the target robot data. Crucially, rather than assigning all retrieved samples equal importance, ReWeight further estimates their relative cross-embodiment relevance and converts it into continuous sample-level weights during VLA post-training. Highly aligned human data therefore provides stronger supervision, while the influence of less transferable samples is suppressed. In this way, ReWeight moves beyond post-training with only robot data or naive mixture of human-robot data, turning abundant human data into selectively retrieved and sample-specific weighted supervision. 

\par
The contributions of this work are summarized below:
\begin{itemize}
    \item We formulate the use of human data for VLA post-training as a joint retrieval-and-weighting problem, distinguishing between which human demonstrations should be transferred and how strongly each selected sample should influence learning.
    \item We introduce ReWeight, a framework that learns a unified visuomotor representation to retrieve demonstration-level human data and derive sample-level weights according to their relevance to target robot data, enabling selective use of heterogeneous human data during VLA post-training.
\end{itemize}
To evaluate the superiority of ReWeight, we conducted extensive experiments across 8 simulation tasks in the RoboTwin 2.0 benchmark and 4 real-world tasks on the DoBot dual arm platform under the clean and randomized settings. Compared to post-training with only robot data, ReWeight improves the average success rate from 39\% to 57\% in simulation and from 40.0\% to 68.8\% in the real world. These results demonstrate the importance of selective retrieval and fine-grained relevance weighting for effectively exploiting egocentric human data in VLA post-training. 

\par
% The remainder of this work is organized as follows: \textit{Section~\ref{sec:preliminary}} formulates the problem, \textit{Section~\ref{sec:method}} presents the proposed method, and Section~\ref{sec:experiments} describes the simulation and real-world experiments, followed by an analysis of the results. Finally, \textit{Section~\ref{sec:conclusion}} concludes this work.

\begin{figure*}[t!]
    \centering
    \includegraphics[width=1.0\textwidth]{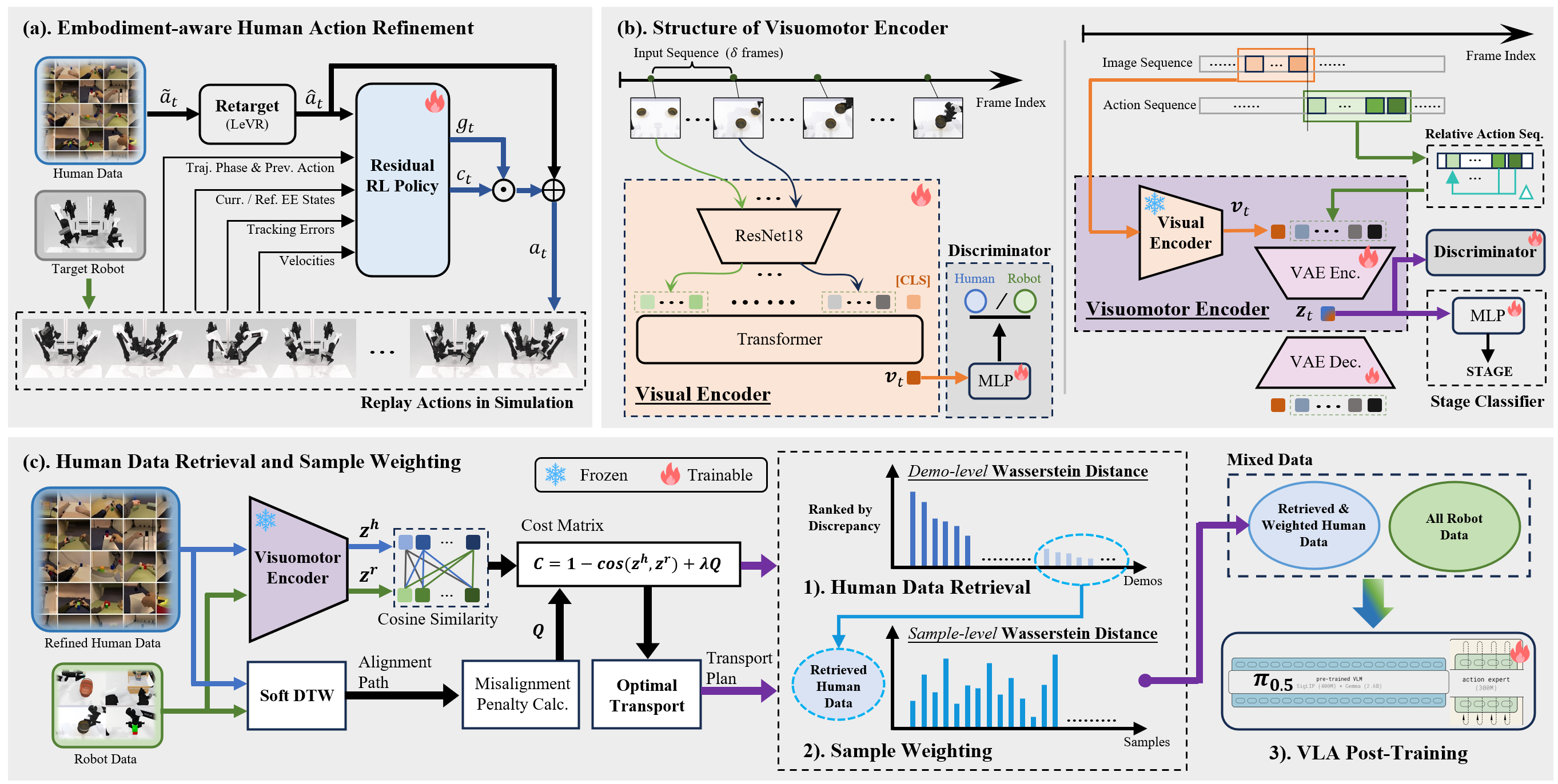}
    \caption{
        Overview of the proposed ReWeight framework. 
}
    \label{fig:method_overview}
\end{figure*}

\section{Preliminary}
\label{sec:preliminary}
\subsection{Problem Description}
\par
Given $M$ robotic demonstrations $\mathcal{D}_r$ and $N$($>M$) human videos $\mathcal{D}_h^{\text{raw}}$, our goal is to select a subset of the human data $\mathcal{S}_h^*$ that is most similar to the target robot data. Such that, post-training the VLA model on the combined mixture data, $\mathcal{D}_{\text{mix}}=\mathcal{S}_h^*\bigcup\mathcal{D}_r$, can enable the resulting policy to maximally leverage relevant human manipulation diversity while boosting downstream task performance. This VLA post-training method is formulated as below:
\begin{equation}
\label{eq:problem_def}
    \mathcal{S}^*_{h}=\arg\underset{\mathcal{S}_h\subseteq\mathcal{D}_{h},|\mathcal{S}_h|=M, N>M}{\max}{\text{Sim}\left(\mathcal{S}_h, \mathcal{D}_r\right)},
\end{equation}
where the datasets are defined as:
\begin{equation*}
\begin{split}
    \mathcal{D}_{r}&=\left\{\mathbf{d}^{r(j)}=\left(\mathbf{o}^r, \mathbf{a}^r, \mathbf{\ell}^r\right)^{(j)}\right\}_{j=1}^{M}, \\
    \mathcal{D}_{h}^{\text{raw}}&=\left\{\mathbf{d}^{h(i)}=\left(\mathbf{o}^h, \Tilde{\mathbf{a}}^h, \mathbf{\ell}^h\right)^{(i)}\right\}_{i=1}^{N}, \\
\end{split}
\end{equation*}
where $\mathbf{d}^{r/h}$ denotes the robot/human demonstration, and the superscript $(\cdot)$ represents the index of demonstrations. $\mathbf{o}^{h/r}$ represents the human/robot visual observations, $\mathbf{a}^r$ denotes the control commands for the robot, $\tilde{\mathbf{a}}^h$ denotes the estimated hand kinematics from human videos, and $\mathbf{\ell}^{h/r}$ is the corresponding textual command of the human/robot data. $\mathcal{S}_h^*=\{(\mathbf{o}_h, \mathbf{a}_h, \mathbf{\ell}_h, \mathbf{w}_h)^{(i)}\}_{i=1}^{N}$ represents an optimal subset extracted from $\mathcal{D}_h^{\text{raw}}$, where each sample is assigned a weight. This subset is constrained to a fixed size $M$ that maximizes a semantic similarity metric $\text{Sim}(\cdot, \cdot)$. 

\subsection{Optimal Transport for Data Discrepancy Measurement}
\par
In this work, to measure the similarity, we use optimal transport (OT) to quantify both the \textit{demonstration-level} discrepancy between human and robot demonstrations, and the \textit{sample-level} discrepancy between the human sample and robot demonstrations. These measures will be used to guide human demonstration retrieval and sample weighting during training.

\par
Given a pair of human and robot demonstrations $\left(\mathbf{d}^h,\mathbf{d}^r\right)$, we quantify their discrepancy using the entropically regularized \textit{Wasserstein distance}, $\mathcal{W}(\cdot)$, defined as:
\begin{equation}
    \label{eq:wasser_distance}
    \begin{split}
        \mathcal{W}\left(\mathbf{d}^h,\mathbf{d}^r,\mathbf{P}\right) &= \sum_{i=1}^{n}\sum_{j=1}^{m}\mathbf{C}_{i,j}\mathbf{P}_{i,j} - \epsilon\mathcal{H}\left(\mathbf{P}\right),  \\
    \end{split}
\end{equation}
where $\mathbf{P}\in\mathbb{R}^{n\times m}$ is the transport plan, and $n$ and $m$ are the number of timesteps in the human and robot demonstrations, respectively. The cost matrix $\mathbf{C}\in\mathbb{R}^{n\times m}$ specifies the  pairwise transport costs, where $\mathbf{C}_{i,j}$ represents the cost of matching the human observation at the $i$-th timestep with the robot observation at the $j$-th timestep. $\mathcal{H}\left(\mathbf{P}\right)$ denotes the entropy of the transport plan, and $\epsilon$ controls the strength of the entropy regularization. The optimal transport plan is obtained by solving Eq. (\ref{eq:ot_problem}) using the \textit{Sinkhorn} algorithm~\cite{cuturi2013sinkhorn}.
\begin{equation}
    \label{eq:ot_problem}
    \begin{split}
        \mathbf{P}^{*}&=\arg\min_{\mathbf{P}\in\mathcal{M}} \mathcal{W}(\mathbf{\mathbf{d}}^h,\mathbf{\mathbf{d}}^r, \mathbf{P}), \\
        \mathcal{M}&=\left\{\mathbf{P}:\mathbf{P}\mathbf{1}_n=\mathbf{1}_n/n,~\mathbf{P}^{T}\mathbf{1}_m=\mathbf{1}_m/m\right\},
    \end{split}
\end{equation}
where $\mathcal{M}$ is the set of admissible transport plans with uniform marginal distributions. 

\par
Given the optimal transport plan, the \textit{demonstration-level} discrepancy $\bar{\mathcal{W}}$ between the human and robot demonstrations and the \textit{sample-level} discrepancy $\bar{\mathcal{W}}_i$ associated with the $i$-th human timestep are quantified, respectively, as below: 
\begin{equation}
    \label{eq:discrepency}
    \bar{\mathcal{W}}\left(\mathbf{d}^h,\mathbf{d}^r\right) = \sum_{i=1}^{n}\mathcal{W}_i,~
    \bar{\mathcal{W}}_i\left(\mathbf{d}_i^h,\mathbf{d}^r\right) = \sum_{j=1}^{m}\mathbf{C}_{i,j}\mathbf{P}^*_{i,j}.
\end{equation}

\section{Method}
\label{sec:method}
\par
This section details ReWeight, a framework for leveraging egocentric human data for VLA post-training. 
% ReWeight treats human demonstrations as data from a distinct embodiment. 
As illustrated in Fig.~\ref{fig:method_overview}, to bridge the human-robot embodiment gap, we refine the extracted hand kinematics and learn a cross-embodiment visuomotor representation for measuring human–robot similarity. The resulting similarity scores are used to retrieve relevant human demonstrations and derive sample-specific weights for VLA post-training. 

\subsection{Embodiment-Aware Human Action Refinement}
\label{sec:execution_refinement}
\par
As shown in Fig. \ref{fig:method_overview} (a), to coarsely align human motion with the robot embodiment, we employ LeVR~\cite{weng2025levr} to map the estimated human hand kinematics to the robotic action space, yielding retargeted trajectory $\hat{\mathbf{a}}^{h}$, where the gripper state is inferred directly from the distance between the human thumb and fingertips. To further resolve the violation of the dynamic constraints such as velocity limits, acceleration bounds, etc., we refine $\hat{\mathbf{a}}^{h}$ with reinforcement learning. We train a residual policy on top of the retargeted trajectory~\cite{johannink2019residual} sampled via the reference state initialization (RSI)~\cite{peng2018deepmimic} technique. At each timestep, the policy predicts both a residual correction $\mathrm{c}_t$ and a non-negative gate $\mathrm{g}_t$ to control the magnitude of the correction.
\begin{equation}
\label{eq:action_refine}
\mathbf{a}_t = \hat{\mathbf{a}}_t + g_t\cdot\mathbf{c}_t,~~g_t = \mathrm{sigmoid}(\mathbf{s}_t) \in (0,1),
\end{equation}
where $\mathbf{s}_t$ denotes the input state including the current and reference end-effector states, tracking errors, velocities, trajectory phase, and previous action. Following the reward design adopted in BeyondMimic~\cite{liao2026beyondmimic}, the per-step reward, combining end-effector pose tracking, velocity matching, and regularization terms, is formulated as:
\begin{equation*}
    r_t = r_{\mathrm{pose}} + r_{\mathrm{vel}} + r_{\mathrm{reg}}.
\end{equation*}

\par
Ultimately, this process yields a refined human dataset with robot-embodiment-aware motion $\mathcal{D}_{h}=\{(\mathbf{o}_h, \mathbf{a}_h, \mathbf{\ell}_h)^{(i)}\}_{i=1}^{N}$. 
% For brevity, in the followings, we refer to the human data as $\mathcal{D}_h$ unless otherwise stated.

\subsection{Cross-Embodiment Visuomotor Encoder}
\label{sec:subsec:visuomoter_encoder}
\par
To obtain a unified representation of the robot and human data for subsequent retrieval and sample weighting, we develop a visuomotor encoder comprising a visual encoder and a visuomotor fusion module, as shown in Fig. \ref{fig:method_overview} (b). 

\subsubsection{Visual Encoder Pretraining}
\label{subsec:visual_encoder}
\par
The visual encoder $\mathbf{E}(\cdot)$ comprises a ResNet-18 backbone and a temporal Transformer. It encodes a sequence of input images $\mathbf{o}_{t-\delta:t}$ into a temporal visual representation $\mathbf{v}_t = \mathbf{E}(\mathbf{o}_{t-\delta:t})$, where $\delta=8$ represents the length of the input image sequence. The encoder is optimized in two stages: it is first warmed up with Eq. (\ref{eq:loss_vis_warmup}) to capture task-relevant visual features, then aligned across human and robot demonstrations via optimizing Eq. (\ref{eq:loss_align}) to obtain a domain-agnostic representation.

% The visual encoder $\mathbf{E}(\cdot)$ comprises a ResNet-18 backbone and a temporal Transformer. It encodes a sequence of input images $\mathbf{o}_{t-\delta:t}$ into a temporal visual representation $\mathbf{v}_t$:
% \begin{equation}
% \label{eq:vis_encoder}
% \mathbf{v}_t = \mathbf{E}(\mathbf{o}_{t-\delta:t})
% \end{equation}
% where $\delta=8$ represents the length of the input image sequence. The encoder is optimized in two stages: it is first warmed up with Eq. (\ref{eq:loss_vis_warmup}) to capture task-relevant visual features, then aligned across human and robot demonstrations via optimizing Eq. (\ref{eq:loss_align}) to obtain a domain-agnostic representation.

\textit{a. warm-up stage}: To capture task-relevant visual features, the visual encoder is trained from scratch using the following warm-up objective:
\begin{equation}
    \label{eq:loss_vis_warmup}
    \mathcal{L}_{\text{warmup}} = \lambda_{\text{sw}}\mathcal{L}_{\text{SwAV}} + \lambda_{\text{tcn}}\mathcal{L}_{\text{tcn}} + \lambda_{\text{tcc}}\mathcal{L}_{\text{tcc}}
\end{equation}
where $\mathcal{L}_{\text{SwAV}}$~\cite{caron2020unsupervised} enforces spatial consistency by contrasting cluster assignments of different augmented views of the same frame; $\mathcal{L}_{\text{tcn}}$~\cite{sermanet2018time} denotes the time-contrastive loss. To align coarse task-relevant dynamics across varying sequence lengths, a temporal cycle-consistency loss $\mathcal{L}_{\text{tcc}}$~\cite{dwibedi2019temporal} is applied. 

\par
\textit{b. alignment stage:} To align visual representations across human and robot embodiments, we extend the warm-up objective with domain-adversarial learning to suppress embodiment-specific visual cues and promote domain-invariant representations, together with an OT-based correspondence loss, which establishes soft cross-embodiment correspondences and brings task-consistent observations closer in the latent space. The overall visual alignment objective is defined as:
\begin{equation}
    \label{eq:loss_align}
    \mathcal{L}_{\text{align}} = \lambda_{\text{tcn}}\mathcal{L}_{\text{tcn}} + \lambda_{\text{tcc}}\mathcal{L}_{\text{tcc}} +  \lambda_{\text{ot}}\mathcal{L}_{\text{ot}}+\lambda_{\text{adv}}\mathcal{L}_{\text{adv}} 
\end{equation}
where $\mathcal{L}_{\text{adv}}$ denotes the adversarial loss. $\mathcal{L}_{\text{ot}}$ represent the OT correspondence loss, defined as the expected discrepancy over sampled human-robot demonstration pairs:
\begin{equation}
\label{eq:loss_ot}
\mathcal{L}_{\mathrm{ot}}=\mathbb{E}_{\mathbf{o}^{h}_{t:t-\delta}\sim\mathcal{D}^{h}}\left[\bar{\mathcal{W}}\left(\mathbf{v}^{h}_t,\mathbf{v}^{r}\right)\right],~~\mathbf{v}^{h/r}_t=\mathbf{E}(\mathbf{o}^{h/r}_{t-\delta:t}).
\end{equation}

\par
To compute $\mathcal{L}_{\text{ot}}$, we first construct a transport cost matrix $\mathbf{C}$ [see Eq. (\ref{eq:modified_cost_matrix})]. With this matrix, we then solve the entropically regularized OT problem in Eq.~(\ref{eq:ot_problem}) to obtain the optimal transport plan $\mathbf{P}^{*}$ and compute the corresponding OT discrepancy $\bar{\mathcal{W}}$ using Eq.~(\ref{eq:discrepency}). Note that, in our formulation, $\mathbf{C}$ combines visual feature similarity with a temporal mismatch penalty $\mathbf{Q}$:
\begin{equation}
    \label{eq:modified_cost_matrix}
    \begin{split}
        \mathbf{C}_{i,j} &= 1 - \text{cos}(\mathbf{v}_i^h, \mathbf{v}_j^r) + \lambda_{\text{dtw}}\mathbf{Q}_{i,j},~~
        \textbf{Q}_{ij} = \frac{|j-p_i|}{(\text{m}-1)},
    \end{split}
\end{equation}
where $\mathbf{v}i^h$ and $\mathbf{v}j^r$ denote the visual features of the $i$-th human frame and the $j$-th robot frame, respectively. The index $p_i$ denotes the robot frame aligned with the $i$-th human frame according to the alignment path estimated by soft-DTW~\cite{cuturi2017soft}. Accordingly, $\mathbf{Q}_{i,j}$ penalizes transport between observations that are far apart from the estimated temporal correspondence.

\subsubsection{Visuomotor Representation Learning}
\par
To account for motion compatibility between human and robot demonstrations, we construct a visuomotor encoder that incorporates motion information into the visual representation. The encoder follows an InfoVAE-based architecture~\cite{zhao2019infovae}, where the visual representation and its corresponding future action chunk are encoded by separate MLPs and subsequently fused into a latent visuomotor representation $\mathbf{z}$. The future action chunk $\mathbf{a}_{t:t+L}$, with $L$ empirically set to 20, is represented relative to its initial pose while preserving the original gripper state.

\par
A stage prediction head, $\hat{\mathrm{y}}=g_{\mathrm{stage}}(\mathbf{z})$, is used to preserve task-stage information, while an auxiliary loss $\mathcal{L}_{\mathrm{aux}}$ promotes cross-embodiment alignment within the same stage and separation across different stages. The latent objective is:
\begin{equation}
\label{eq:loss_latent}
\mathcal{L}_{\mathrm{lat}}=\mathcal{L}_{\mathrm{ce}}+\lambda_{\mathrm{aux}}\mathcal{L}_{\mathrm{aux}}
\end{equation}
where $\mathcal{L}_{ce}=\mathrm{CE}(\mathrm{y}, \hat{\mathrm{y}})$ denotes the cross entropy loss, and
\begin{equation*}
    \begin{split}
        \mathcal{L}_{\mathrm{aux}} &= \mathbb{E}\left[-\mathrm{cos}(\mathbf{z}^{h}_s,\mathbf{z}^{r}_s)\right] + \mathbb{E}\left[\max(0, \mathrm{cos}(\mathbf{z}_{s}, \mathbf{z}_{s^\prime})-\mu)\right],
    \end{split}
\end{equation*}
where $\mathbf{z}^{h}_{s}$ and $\mathbf{z}^{r}_{s}$ denote latent representations from human and robot demonstrations at the same stage $s$, respectively, while $\mathbf{z}_{s}$ and $\mathbf{z}_{s^\prime}$ denote representations from different stages, with $s\neq s'$. The first term maximizes the similarity between same-stage cross-domain representations, whereas the second penalizes the similarity between different-stage representations when it exceeds the margin $\mu=0.3$. The overall VAE training objective is formulated as: 
\begin{equation}
\label{eq:loss_vae}
\mathcal{L}_{\mathrm{vae}} = 
\lambda_{\mathrm{rec}}\mathcal{L}_{\mathrm{rec}}
+\lambda_{\mathrm{reg}}\mathcal{L}_{\mathrm{reg}}
+\lambda_{\mathrm{lat}}\mathcal{L}_{\mathrm{lat}},
\end{equation}
where $\mathcal{L}_{\mathrm{rec}}$ is the reconstruction loss for the visual feature and relative motion trajectory. The regularization term $\mathcal{L}_{\mathrm{reg}}$ follows InfoVAE~\cite{zhao2019infovae} combining the KL divergence and maximum mean discrepancy. The latent objective $\mathcal{L}_{\mathrm{lat}}$ further enforces the latent space according to demonstration stage while reducing stage-level discrepancies between human and robot data. These objectives encourage the latent representation to retain the visual and motion information required for reconstruction, suppress embodiment-specific variations, and preserve the temporal progression of the manipulation task.

\subsection{Human Demonstration Retrieval and Sample Weighting}
\label{sec:retrieval_weighting}
\par
We employ a two-level strategy for \textit{demonstration-level} retrieval and \textit{sample-level} weighting of the human data for VLA post-training.

\subsubsection{Demonstration-level Retrieval}
\par
For each candidate human demonstration $\mathbf{d}^{h(i)}$, we compute its discrepancy from every robot demonstration of the same task. Let
\begin{equation*}
\mathbf{s}^{(i)}=\left[s_1^{(i)},\ldots,s_M^{(i)}\right]\in\mathbb{R}^{M},
~s_j^{(i)} = \bar{\mathcal{W}}\left(\mathbf{d}^{h(i)},\mathbf{d}^{r(j)}\right),
\end{equation*}
where $M$ is the number of robot demonstrations in each task, and $\bar{\mathcal{W}}(\cdot,\cdot)$ is the demonstration-level discrepancy defined in Eq.~(\ref{eq:discrepency}). When constructing the cost matrix in Eq.~(\ref{eq:modified_cost_matrix}), we compute the cosine distance using the latent features $\mathrm{z}$ produced by the visuomotor encoder developed in \emph{Sec.~\ref{sec:subsec:visuomoter_encoder}}. We sort the elements of $\mathbf{s}^{(i)}$ in ascending order as
$s_{(1)}^{(i)}\leq\cdots\leq s_{(M)}^{(i)}$
and define the retrieval score of $\mathbf{d}^{h(i)}$ as the average of its $K$ smallest discrepancies: $\bar{s}^{(i)}=\sum_{k=1}^{K}s_{(k)}^{(i)}/K$.
We empirically set $K=5$ in all experiments as in~\cite{haldar2023watch}. After ranking the human demonstrations by $\bar{s}^{(i)}$, we select the $M$ lowest-scoring ones to construct $\mathcal{S}_h^*$, yielding a $1{:}1$ human-to-robot data ratio for VLA post-training as in~\cite{kareer2025emergence}.

\subsubsection{Sample-level Weighting}
\par
After retrieval, individual frames within the retained demonstrations may still differ in their relevance to the robot data. We thereby assign a discrepancy-aware weight to each retained human sample. For a human sample at timestep $t$ in $\mathbf{d}^{h(i)}\in\mathcal{S}_h^{*}$, its sample-level discrepancy is defined as:
\begin{equation}
\label{eq:sample_discrepancy}
d_t^{(i)} = 
\sum_{j\in\mathcal{N}_K^{(i)}}
\mathcal{W}_t \left(\mathbf{d}^{h(i)}_t,\mathbf{d}^{r(j)}\right)/K,
\end{equation}
where $\mathcal{N}_K^{(i)}$ denote the indices of its $K$ nearest robot demonstrations identified during retrieval. $\mathcal{W}_t(\cdot,\cdot)$ denotes the frame-level discrepancy defined in Eq.~(\ref{eq:discrepency}). A smaller value of $d_t^{(i)}$ indicates that the human sample is more consistent with the corresponding robot demonstrations.

\par
To assign larger weights to samples with smaller cross-embodiment discrepancies, we first convert the discrepancy into an unnormalized similarity weight:
\begin{equation}
\label{eq:uncalibrated_weight}
\widetilde{w}_t^{(i)}=d_{\max}-d_t^{(i)},~~
d_{\max}=\max_{(u,l)\in\mathcal{I}_h^{*}}d_l^{(u)},
\end{equation}
where and $\mathcal{I}_h^{*}$ means the set of all samples contained in the retrieved human demonstrations. The normalized weight is then computed as:
% \begin{equation}
% \label{eq:calibrated_weight}
% w_t^{(i)} = \alpha + (1-\alpha)\frac{\widetilde{w}_t^{(i)}}{\sum_{(u,l)\in\mathcal{I}_h^{*}}\widetilde{w}_l^{(u)}},
% \end{equation}
\begin{equation}
\label{eq:calibrated_weight}
% w_t^{(i)} = \alpha + (1-\alpha)\frac{\widetilde{w}_t^{(i)}}{\displaystyle
% \max_{(u,l)\in\mathcal{I}_h^{*}}\widetilde{w}_l^{(u)}},
w_t^{(i)} = \alpha + (1-\alpha)\cdot\widetilde{w}_t^{(i)}/\displaystyle
\max_{(u,~l)\in\mathcal{I}_h^{*}}\widetilde{w}_l^{(u)},
\end{equation}
where $\alpha\in[0,1]$ is a sample-weight calibration parameter, that ensures sufficient influence of human samples.

\subsubsection{Loss for Post-Training}
\par
During post-training, the model is optimized using the mixed robotic data $\mathcal{D}_r$ and the retrieved human data from $\mathcal{S}_h^*$. Let $\ell_i(\theta)$ denote the per-sample VLA loss for sample $i$. The post-training loss is defined as:

% During VLA post-training, we jointly optimize the model using robot data $\mathcal{D}_r$ and the retrieved human data from $\mathcal{S}_h^*$. Let $\ell_i(\theta)$ denote the per-sample VLA loss for sample $i$. The post-training objective is defined as:
\begin{equation}
\label{eq:loss_vla}
% \mathcal{L}_{\mathrm{VLA}}=\frac{\sum_{i\in\mathcal{B}_h}w_i\ell_i(\theta)}{\sum_{i\in\mathcal{B}_h}w_i}
% +\frac{\sum_{j\in\mathcal{B}_r}\ell_j(\theta)}{\left|\mathcal{B}_r\right|}
% \mathcal{L}_{\mathrm{VLA}}=\frac{\sum_{i\in\mathcal{B}_r}\ell_i(\theta)+\sum_{i\in\mathcal{B}_h}w_j\ell_j(\theta)}{\left|\mathcal{B}\right|}
\mathcal{L}_{\mathrm{VLA}}=\left(\sum_{i\in\mathcal{D}_r}\ell_i(\theta)+\sum_{i\in\mathcal{S}_h^*}w_j\ell_j(\theta)\right)/\left|\mathcal{D}_{\mathrm{mix}}\right|
\end{equation}
where $\mathcal{D}_{\mathrm{mix}}=\mathcal{D}_r\cup\mathcal{S}_h^*$. $\left|\mathcal{D}_{\mathrm{mix}}\right|$ denotes the number of samples. The human sample weight $w_j$ is computed by Eq. (\ref{eq:calibrated_weight}). 

\subsection{Implementation Details}
\label{subsec:implementation_details}
\subsubsection{Residual RL Policy}
% simulator / input-output / RL algorithm / steps / GPU
The residual RL policy for human action refinement adopts an actor--critic architecture and is optimized with PPO~\cite{schulman2017proximal}. We use a learning rate of $1\times10^{-4}$, a batch size of 64, and a discount factor of 0.95. For each task, we train a separate residual policy for 50k environment steps, using RoboTwin 2.0 for the AgileX platform and MuJoCo for the DoBot platform. After training, the policy is frozen and applied to refine all retargeted human trajectories before visuomotor representation learning. 

\subsubsection{Visuomotor Encoder}
\par
We randomly sample human demonstrations to match the number of available robot demonstrations for each task, and combine the sampled human data with all robot demonstrations for training. The visuomotor encoder is optimized with AdamW using a learning rate of $1\times10^{-4}$ and a batch size of 64. Training proceeds in three stages: the visual encoder is warmed up for 20 epochs using Eq.~(\ref{eq:loss_vis_warmup}), further aligned across human and robot domains for 260 epochs using Eq.~(\ref{eq:loss_align}), and the visuomotor encoder is then trained for 500 epochs with the latent objective in Eq.~(\ref{eq:loss_vae}). All training is performed on a single NVIDIA RTX 5090 GPU.
% \todo{do you have a time to show?}

\subsubsection{VLA Post-Training}
\par
We adopt $\pi_{0.5}$~\cite{pmlr-v305-black25a} as the VLA backbone and initialize all models from the same pretrained checkpoint. To ensure a fair comparison, we use identical optimization settings across all methods. Each model is post-trained with AdamW on two NVIDIA H100 GPUs using a batch size of 64 and a learning rate of $2.5\times10^{-5}$. We use only the external camera view for policy training and deployment. Models are trained for 20k optimization steps in simulation and 40k steps in the real-world experiments.

\begin{figure}[t]
    \centering
    \includegraphics[width=\columnwidth]{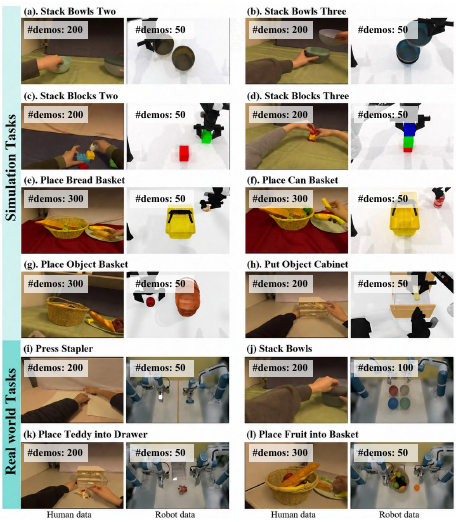}
    \caption{Representative tasks in simulation and real-world settings, with the number of collected demonstrations indicated for each task. 
    % \todo{change the color of "demo" in the figure, make it more visiable}
    % The number of demonstrations for each task is indicated in the upper-left corner of each image.
    }
    \label{fig:task_overview}
\end{figure}

\begin{table*}[!t]
\centering
\caption{
Comparison of task success rates for models trained with different robot-human data composition strategies \textit{(refer to Sec.~\ref{subsubcsec:comparison_study} for details)} under clean and randomized simulation settings. Success rates are computed over 100 trials per task.
}
\label{tab:main_retrieval_weighting}
\setlength{\tabcolsep}{8pt}
\resizebox{\textwidth}{!}{
\begin{tabular}{lccccccc}
\toprule
\multirow{2}{*}{\textbf{Tasks}} 
& \multirow{2}{*}{\textbf{Robot-Only}} 
& \multicolumn{6}{c}{\textbf{Mixing Data} \scriptsize{(All Robot Data + Retrieved Human Data)}} \\
\cmidrule(lr){3-8}
& & All$_{1.0}$ & Random$_{1.0}$ & \multicolumn{1}{|c}{Retrieval$_{0.4}$} & Retrieval$_{0.7}$ & Retrieval$_{1.0}$ & \multicolumn{1}{|c}{\textbf{ReWeight} (ours)} \\
\midrule
\multicolumn{8}{l}{\textbf{(a). Clean settings}} \\
Stack Bowls Two      & 0.88 & 0.87 & 0.88 & 0.89 & 0.87 & \underline{0.95} & \textbf{0.96} \\
Stack Bowls Three    & 0.57 & 0.62 & 0.63 & 0.61 & 0.67 & \underline{0.71} & \textbf{0.78} \\
Stack Blocks Two     & 0.30 & 0.51 & 0.44 & 0.47 & 0.68 & \underline{0.71} & \textbf{0.74} \\
Stack Blocks Three   & 0.14 & 0.16 & 0.22 & 0.26 & 0.24 & \underline{0.29} & \textbf{0.30} \\
Place Bread Basket   & 0.19 & 0.20 & 0.22 & 0.17 & 0.17 & \underline{0.25} & \textbf{0.31} \\
Place Can Basket     & 0.29 & 0.30 & 0.31 & 0.29 & \underline{0.33} & 0.32 & \textbf{0.37} \\
Place Object Basket  & 0.34 & 0.44 & 0.38 & 0.47 & \underline{0.48} & 0.45 & \textbf{0.51} \\
Put Object Cabinet   & 0.42 & 0.37 & 0.44 & 0.44 & 0.40 & \underline{0.51} & \textbf{0.55} \\
\textbf{Average}     & 0.39 & 0.43 & 0.44 & 0.45 & 0.48 & \underline{0.52} & \textbf{0.57} \\
\midrule
\multicolumn{8}{l}{\textbf{(b). Randomized settings}}\\
Stack Bowls Two      & 0.58 & 0.55 & 0.61 & 0.61 & 0.57 & \underline{0.65} & \textbf{0.68} \\
Stack Bowls Three    & \textbf{0.48} & 0.34 & 0.41 & 0.46 & \textbf{0.48} & \underline{0.47} & \textbf{0.48} \\
Stack Blocks Two     & 0.19 & 0.10 & 0.12 & 0.27 & \underline{0.28} & 0.25 & \textbf{0.29} \\
Put Object Cabinet   & 0.12 & 0.15 & 0.13 & 0.17 & 0.19 & \underline{0.22} & \textbf{0.23} \\
\textbf{Average}     & 0.34 & 0.29 & 0.32 & 0.38 & 0.38 & \underline{0.40} & \textbf{0.42} \\
\bottomrule
\multicolumn{8}{l}{\scriptsize{Robot-Only: only use robot data; \textbf{Best} and \underline{second-best} results are shown in bold and underlined, respectively.}} \\
% \multicolumn{8}{l}{\scriptsize{Each success rate is computed from 100 evaluation trials.}} \\
\end{tabular}}
% \vspace{-0.8em}
\end{table*}

\section{Experiments and Results}
\label{sec:experiments}
\par
To evaluate the effectiveness of ReWeight, we conduct experiments in both simulation and real-world environments. We compare ReWeight with robot-only training and naive human-robot data mixing, assessing VLA policy performance under clean and randomized settings across eight simulated and four real-world tasks.

% To evaluate the effectiveness of the proposed ReWeight, we conduct experiments in both simulation and real-world environments. we compares ReWeight against robot-only training and naive human-robot data mixing method to assess its impact on VLA policy performance under the clean and randomized settings across eight simulation and four real-world tasks. 
% In addition, we conducted ablation studies on the key design choices and hyperparameters.

\subsection{Simulation Settings}
\par
RoboTwin 2.0~\cite{chen2025robotwin} benchmark has been used to evaluate ReWeight across eight bimanual manipulation tasks, including four stacking, three pick-and-place, and one drawer-manipulation. Fig.~\ref{fig:task_overview} shows robot data and the corresponding egocentric human data from EgoDex~\cite{hoque2026egodex}, together with their data sizes. We divide the eight tasks into two groups based on their manipulation characteristics: the four stacking tasks and the remaining four tasks. Following the training procedure in \textit{Section~\ref{subsec:implementation_details}}, we then train a separate visuomotor encoder for each group to retrieve and weight human data for the corresponding robot tasks. A separate $\pi_{0.5}$~\cite{pmlr-v305-black25a} model is then post-trained for each group using the combined robot and retrieved human data. Unless otherwise stated, in all the following experiments, we use all available robot data and retrieve an equal size of human data for each task.

\begin{table*}[t]
\centering
% \caption{Ablation study of cross-embodiment representation learning.}
\caption{Cross-embodiment alignment performance of visuomotor encoder under different ablation settings.}
\label{tab:visuomotor_ablation}
\resizebox{\textwidth}{!}{
\begin{tabular}{lcccccccccc}
\toprule
% \multicolumn{11}{l}{\textbf{(a) Task Group I}} \\[-1pt]
\multirow{2}{*}{\textbf{Variants}}
& \multicolumn{2}{c}{\textbf{Stack Bowls Two}}
& \multicolumn{2}{c}{\textbf{Stack Bowls Three}}
& \multicolumn{2}{c}{\textbf{Stack Blocks Two}}
& \multicolumn{2}{c}{\textbf{Stack Blocks Three}}
& \multicolumn{2}{c}{\textbf{Average}} \\
\cmidrule(lr){2-3}\cmidrule(lr){4-5}\cmidrule(lr){6-7}\cmidrule(lr){8-9}\cmidrule(lr){10-11}
& \scriptsize{Stage Acc.} & \scriptsize{KF Dist.}
& \scriptsize{Stage Acc.} & \scriptsize{KF Dist.}
& \scriptsize{Stage Acc.} & \scriptsize{KF Dist.}
& \scriptsize{Stage Acc.} & \scriptsize{KF Dist.}
& \scriptsize{Stage Acc.} $\uparrow$ & \scriptsize{KF Dist.} $\downarrow$ \\
\midrule
w/o Action Info. & 0.84 & 11.2 & 0.84 & 5.9 & 0.80 & 11.2 & 0.85 & 11.6 & 0.83 & 10.0 \\
w/o Temporal Penalty & 0.86 & 8.5 & 0.86 & 4.6 & 0.84 & 9.2 & 0.88 & 9.5 & 0.86 & 8.0 \\
Visuomotor Enc. \scriptsize{(Ours)}
& \textbf{0.95} & \textbf{0.8}
& \textbf{0.93} & \textbf{2.6}
& \textbf{0.91} & \textbf{5.6}
& \textbf{0.91} & \textbf{7.5}
& \textbf{0.93} & \textbf{4.1} \\
% \midrule
% \multicolumn{11}{l}{\textbf{(b) Task Group II}} \\[-1pt]
% \multirow{2}{*}{\textbf{Variants}}
% & \multicolumn{2}{c}{\textbf{Place Bread Basket}}
% & \multicolumn{2}{c}{\textbf{Place Can Basket}}
% & \multicolumn{2}{c}{\textbf{Place Object Basket}}
% & \multicolumn{2}{c}{\textbf{Put Object Cabinet}}
% & \multicolumn{2}{c}{\textbf{Average}} \\
% \cmidrule(lr){2-3}\cmidrule(lr){4-5}\cmidrule(lr){6-7}\cmidrule(lr){8-9}\cmidrule(lr){10-11}
% & \scriptsize{Stage Acc.} $\uparrow$ & \scriptsize{KF Dist.} $\downarrow$
% & \scriptsize{Stage Acc.} $\uparrow$ & \scriptsize{KF Dist.} $\downarrow$
% & \scriptsize{Stage Acc.} $\uparrow$ & \scriptsize{KF Dist.} $\downarrow$
% & \scriptsize{Stage Acc.} $\uparrow$ & \scriptsize{KF Dist.} $\downarrow$
% & \scriptsize{Stage Acc.} $\uparrow$ & \scriptsize{KF Dist.} $\downarrow$ \\
% \midrule
% w/o Action Info. & 0.83 & 19.8 & 0.83 & 7.7 & 0.81 & 16.4 & 0.82 & 15.9 & 0.82 & 15.0 \\
% w/o Temporal Penalty & 0.84 & 13.8 & 0.85 & 6.6 & 0.84 & 12.5 & 0.82 & 8.8 & 0.84 & 10.4 \\
% Visuomotor Enc. \scriptsize{(Ours)}
% & \textbf{0.88} & \textbf{9.2}
% & \textbf{0.88} & \textbf{4.3}
% & \textbf{0.89} & \textbf{9.3}
% & \textbf{0.93} & \textbf{7.3}
% & \textbf{0.90} & \textbf{7.5} \\
\bottomrule
\multicolumn{11}{l}{\scriptsize ``w/o Action Info.'' refers to Visual Encoder; ``w/o Temporal Penalty'' refers to without temporal mismatch penalty $\mathbf{Q}$ in Eq.~(\ref{eq:modified_cost_matrix}).} \\
\multicolumn{11}{l}{\scriptsize Stage Acc.: Stage accuracy; ~~~KF Dist.: Key frame distance;~~~Visuomotor Enc.: Visuomotor Encoder.} 
\end{tabular}}
\end{table*}

\subsection{Simulation Results}
\subsubsection{Comparison Study} 
\label{subsubcsec:comparison_study}
\par
Table~\ref{tab:main_retrieval_weighting} compares different strategies for incorporating human data into $\pi_{0.5}$ post-training. To validate the effectiveness of incorporating the human data, we compared post-training with robot data alone (``\textit{Robot Only}") against post-training with a mixture of robot and human data (``\textit{Mixing Data}"). To assess the necessity of data retrieval, we further compared post-training $\pi_{0.5}$ using all available human data (``\textit{All$_{1.0}$}") with using an equally sized, randomly sampled human data (``\textit{Random$_{1.0}$}"); all human samples in these two variants were assigned a fixed weight of 1.0. To evaluate the benefit of sample weighting, we used ReWeight to retrieve human data with high similarity and post-trained the model with uniform sample weights of 0.4, 0.7, and 1.0, denoted as ``\textit{Retrieval$_{0.4}$}", ``\textit{Retrieval$_{0.7}$}", and ``\textit{Retrieval$_{1.0}$}", respectively.
% All models post-trained using the above human-robot data composition strategies were evaluated under the clean and randomized settings of RoboTwin 2.0.

\par
Simply incorporating human data improves performance under the clean setting where ``\textit{All$_{1.0}$}" and ``\textit{Random$_{1.0}$}" increase the average success rate over ``\textit{Robot-Only}" by $4\%$ and $5\%$, respectively. However, these gains do not transfer to the randomized setting. Under the randomized setting, they instead reduce performance by 5\% and 2\% percentage points. This contrast suggests that indiscriminately mixing human and robot data may introduce irrelevant or mismatched samples, leading the model to overfit to the clean training distribution rather than generalize to unseen environmental variations. These findings support our hypothesis that more human data will not automatically result into better policy. The relevance of the selected demonstrations is critical.

\par
The variants trained with human data selected by the proposed demonstration-level retrieval method consistently outperform both ``\textit{All$_{1.0}$}" and ``\textit{Random${1.0}$}" under both clean and randomized settings. In the randomized settings, ``\textit{Retrieval$_{0.4}$}" outperforms ``\textit{Random${1.0}$}" by 6\%, demonstrating the effectiveness of the proposed retrieval method. Among the retrieval variants with uniform weights, increasing the human-data weight from 0.4 (``\textit{Retrieval$_{0.4}$}") to 1.0 (``\textit{Retrieval$_{1.0}$}") improves the average success rate by 7\% in the clean setting and 2\% in the randomized setting. ReWeight further assigns a sample-specific weight to each retrieved sample according to Eq.~(\ref{eq:calibrated_weight}). Compared with the strongest uniform-weight variant, ``\textit{Retrieval${1.0}$}", ReWeight achieves further gains of 5\% and 2\% in the clean and randomized settings, respectively. Overall, these results demonstrate the benefits of combining demonstration-level retrieval with sample-level weighting.

\begin{figure}[t]
    \centering
    \includegraphics[width=\columnwidth]{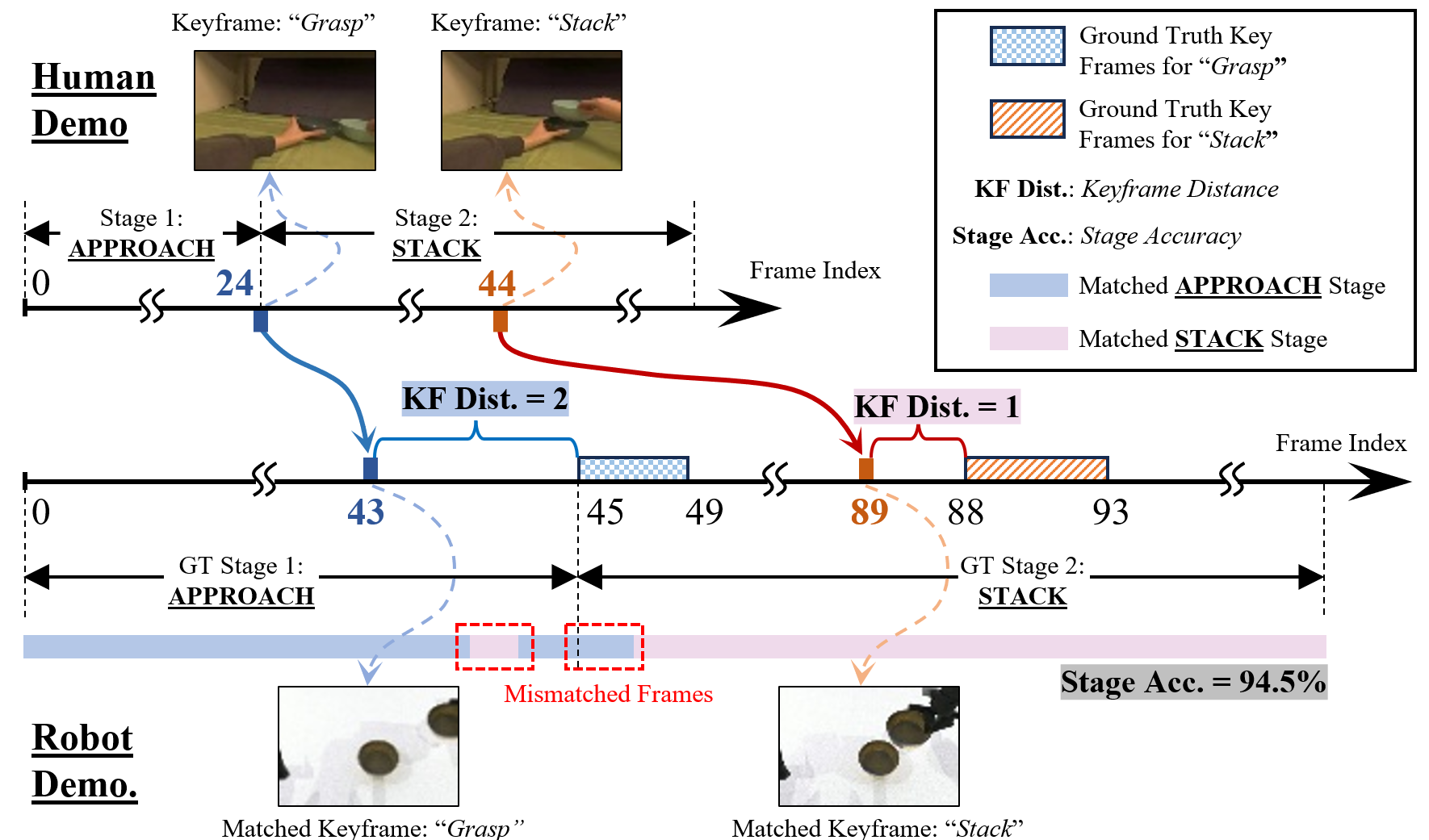}
    \caption{
        Illustration of cross-embodiment key frame correspondence and the calculation of Stage Accuracy (\textit{Stage Acc.}) and Keyframe Distance (\textit{KF Dist.}).
    }
    \label{fig:metrics}
\end{figure}

\begin{figure}[t]
    \centering
    \includegraphics[width=\columnwidth]{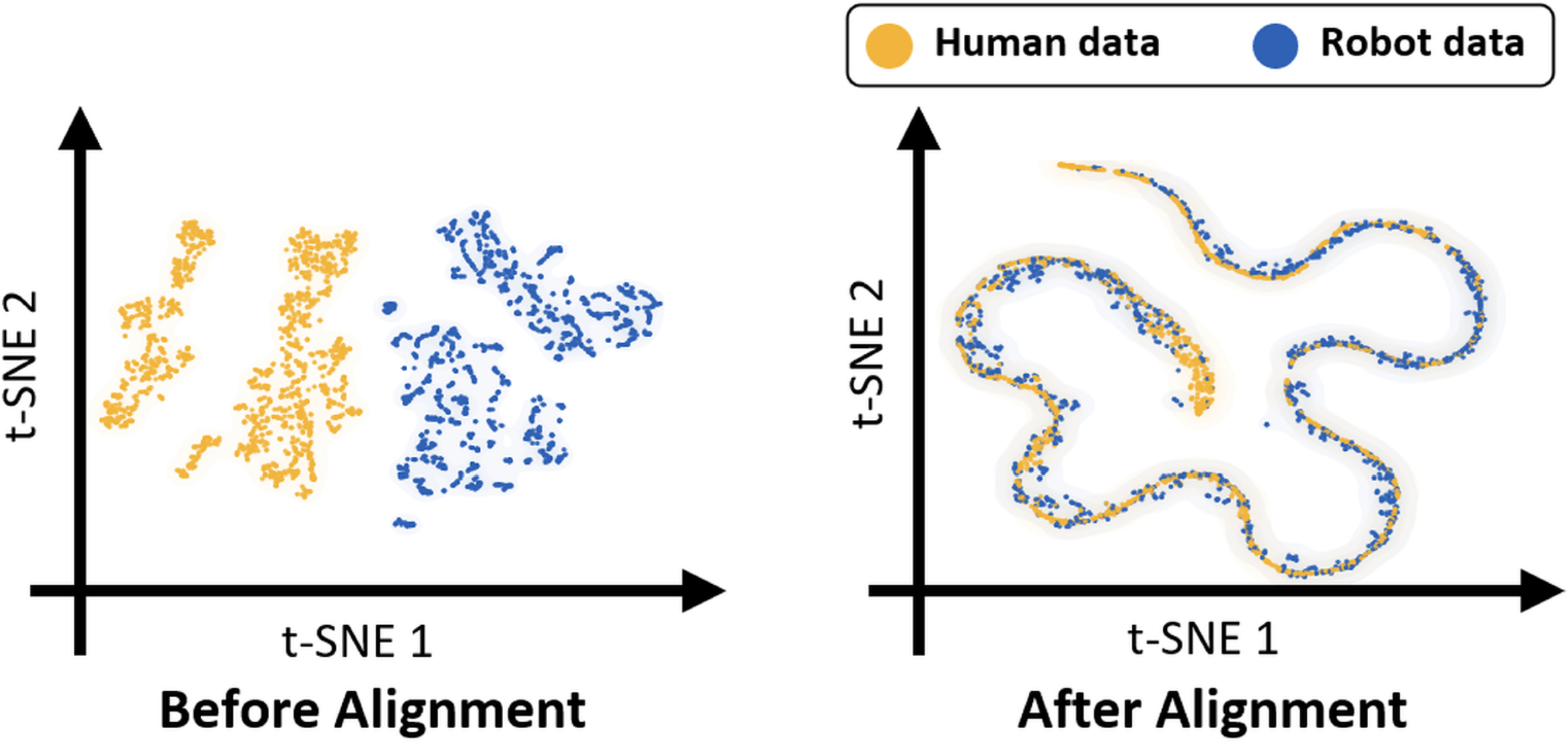}
    \caption{
        t-SNE visualization of human and robot visual representations before and after cross-embodiment alignment.
    }
    \label{fig:visual_tsne}
\end{figure}

\subsubsection{Ablation Study}
\par
% todo: showing the results of the evaluation metrics, stage and the so-called frame distance in figures, with a double column figure, vividly explain what this two matrics are defined and how are they computed.
We conducted ablation studies on two key design choices of the visuomotor encoder: (i) incorporating future action chunks into the learned representation and (ii) imposing a temporal mismatch penalty when solving the OT problem for learning in Eq.~(\ref{eq:loss_ot}) and retrieval with Eq.~(\ref{eq:discrepency}). The results are reported in Table~\ref{tab:visuomotor_ablation}. In ``\textit{w/o Action Info.},'' we remove the action input and rely solely on the learned visual features for cross-embodiment matching. To assess the effectiveness of the temporal penalty $\mathbf{Q}$ in the cost matrix defined in Eq.~(\ref{eq:modified_cost_matrix}), we additionally evaluate a variant without $\mathbf{Q}$ (``\textit{w/o Temporal Penalty}"). We evaluate the encoder variants on four simulation tasks: \textit{Stack Bowls Two}, \textit{Stack Bowls Three}, \textit{Stack Blocks Two}, and \textit{Stack Blocks Three}. Cross-embodiment correspondence quality is assessed using two metrics: Stage Accuracy and Keyframe Distance. Fig.~\ref{fig:metrics} illustrates a representative key frame correspondence between a human demonstration and its matched robot demonstration, together with the definitions of both metrics. Specifically, for each human demonstration, we computed its Wasserstein distance to every robot demonstration from the same task and selected the robot demonstration with the smallest distance as its match. Stage accuracy (\textit{Stage Acc.}) and key frame distance (\textit{KF Dist.}) are then computed for each matched human-robot pair. The values reported in Table~\ref{tab:visuomotor_ablation} are averaged over all human demonstrations across the four tasks.

\begin{table}[t]
\centering
\caption{Success rates under different values of the sample-weight calibration parameter $\alpha$ in Eq.~(\ref{eq:calibrated_weight}).}
\label{tab:alpha_sensitivity}
\setlength{\tabcolsep}{4pt}
\resizebox{\columnwidth}{!}{
\begin{tabular}{lccccc}
\toprule
\multirow{2}{*}{\textbf{Values}}
& \shortstack{\textbf{Stack}\\\textbf{Bowls Three}}
& \shortstack{\textbf{Stack}\\\textbf{Blocks Two}}
& \shortstack{\textbf{Place}\\\textbf{Object Basket}}
& \textbf{Average} \\
\midrule
Robot Only & 0.57 & 0.30 & 0.34 & 0.40  \\
$\alpha=0.1$ & 0.77 & 0.71 & 0.46 & 0.65  \\
$\alpha=0.3$ & 0.77 & 0.70 & 0.47 & 0.65  \\
$\alpha=0.5$ & \textbf{0.78} & \textbf{0.74} & \textbf{0.51} & \textbf{0.68}  \\
$\alpha=0.7$ & 0.74 & 0.72 & 0.46 & 0.64  \\
$\alpha=1.0$ & 0.71 & 0.71 & 0.45 & 0.62  \\
\bottomrule
\multicolumn{5}{l}{\scriptsize Robot Only: no human data for training; $\alpha=1.0$: unweighted human data.} \\
% \multicolumn{5}{l}{\scriptsize $\alpha=1.0$: unweighted human data.}
\end{tabular}
}
\end{table}

\begin{table*}[t]
\centering
\caption{Performance comparison under real-world perturbations, including lighting changes and visual distractors.}
\label{tab:real_world_generalization}
\setlength{\tabcolsep}{12pt}
\resizebox{\textwidth}{!}{
\begin{tabular}{lcccccc}
\toprule
\multirow[c]{2}{*}{\textbf{Tasks}}
& \multicolumn{2}{c}{\textbf{Robot Only}}
& \multicolumn{2}{c}{\textbf{Naive Mixing}}
& \multicolumn{2}{c}{\textbf{ReWeight} (Ours)} \\
\cmidrule(lr){2-3}
\cmidrule(lr){4-5}
\cmidrule(lr){6-7}
& \scriptsize{Lighting} 
& \scriptsize{Distractors} 
& \scriptsize{Lighting} 
& \scriptsize{Distractors} 
& \scriptsize{Lighting} 
& \scriptsize{Distractors} \\
\midrule
Stack Bowls
& 2/10 & 2/10
& 4/10 & 2/10
& \textbf{7/10} & \textbf{5/10} \\
Place Teddy Drawer
& 3/10 & \textbf{5/10}
& \textbf{6/10} & 4/10
& 5/10 & \textbf{5/10} \\
Pick Fruits Basket
& 5/10 & 4/10
& 5/10 & 7/10
& \textbf{7/10} & \textbf{9/10} \\
Press Stapler
& 2/10 & 0/10
& 3/10 & 2/10
& \textbf{5/10} & \textbf{5/10} \\
\midrule
\textbf{Overall}
& 12/40 \scriptsize{(30\%)} & 11/40 \scriptsize{(28\%)}
& 18/40 \scriptsize{(45\%)} & 15/40 \scriptsize{(38\%)}
& \textbf{24/40 \scriptsize{(60\%)}} & \textbf{24/40 \scriptsize{(60\%)}} \\
\bottomrule
\multicolumn{7}{l}{\scriptsize{Robot-Only: only use robot data;~~Naive Mixing: randomly sampled human data + all robot data;~~\textbf{Best} results are shown in bold.}} \\
% \multicolumn{7}{l}{\scriptsize{Each success rate is computed from 100 evaluation trials.}} \\
\end{tabular}
}
\end{table*}

\par
Notably, removing the action input (``\textit{w/o Action Info.}") consistently degrades performance across all four simulation tasks, reducing the average stage accuracy from 0.93 to 0.83 and increasing the average keyframe distance from 4.1 to 10.0 frames. Fig.~\ref{fig:visual_tsne} visualizes the features learned by the vision encoder before and after cross-embodiment alignment using t-SNE. Before alignment, the human and robot features form clearly separated domains, whereas after alignment, they are better integrated into a shared feature space. This confirms the effectiveness of the visual alignment strategy for cross-embodiment similarity estimation. However, the inferior performance of the visual-only variant shows that visual alignment alone is insufficient for reliable retrieval, highlighting the critical role of future action information. Removing the temporal penalty (``\textit{w/o Temporal Penalty}") also consistently degrades both metrics, although to a lesser extent, with a 7\% drop in stage accuracy and a 3.9-frame increase in keyframe distance. These results show that action information is essential for capturing behaviorally meaningful cross-embodiment discrepancies, while the temporal prior promotes temporally consistent human-robot correspondences.

\par
We further investigate the sensitivity of ReWeight to the minimum human-sample weight $\alpha$ in Eq.~(\ref{eq:calibrated_weight}) by evaluating task success rates on three representative simulation tasks, including \textit{Stacking Bowls}, \textit{Stacking Blocks}, and \textit{Placing Object Basket}. A smaller $\alpha$ more strongly suppresses human samples with large cross-embodiment discrepancies, whereas a larger $\alpha$ allows all retrieved samples to contribute more to policy optimization. As shown in Table~\ref{tab:alpha_sensitivity}, $\alpha=0.5$ achieves the best downstream performance by maintaining a reasonable minimum contribution from each retrieved human sample while still differentiating samples according to their cross-embodiment relevance.

\begin{figure}[t]
    \centering
    \includegraphics[width=\columnwidth]{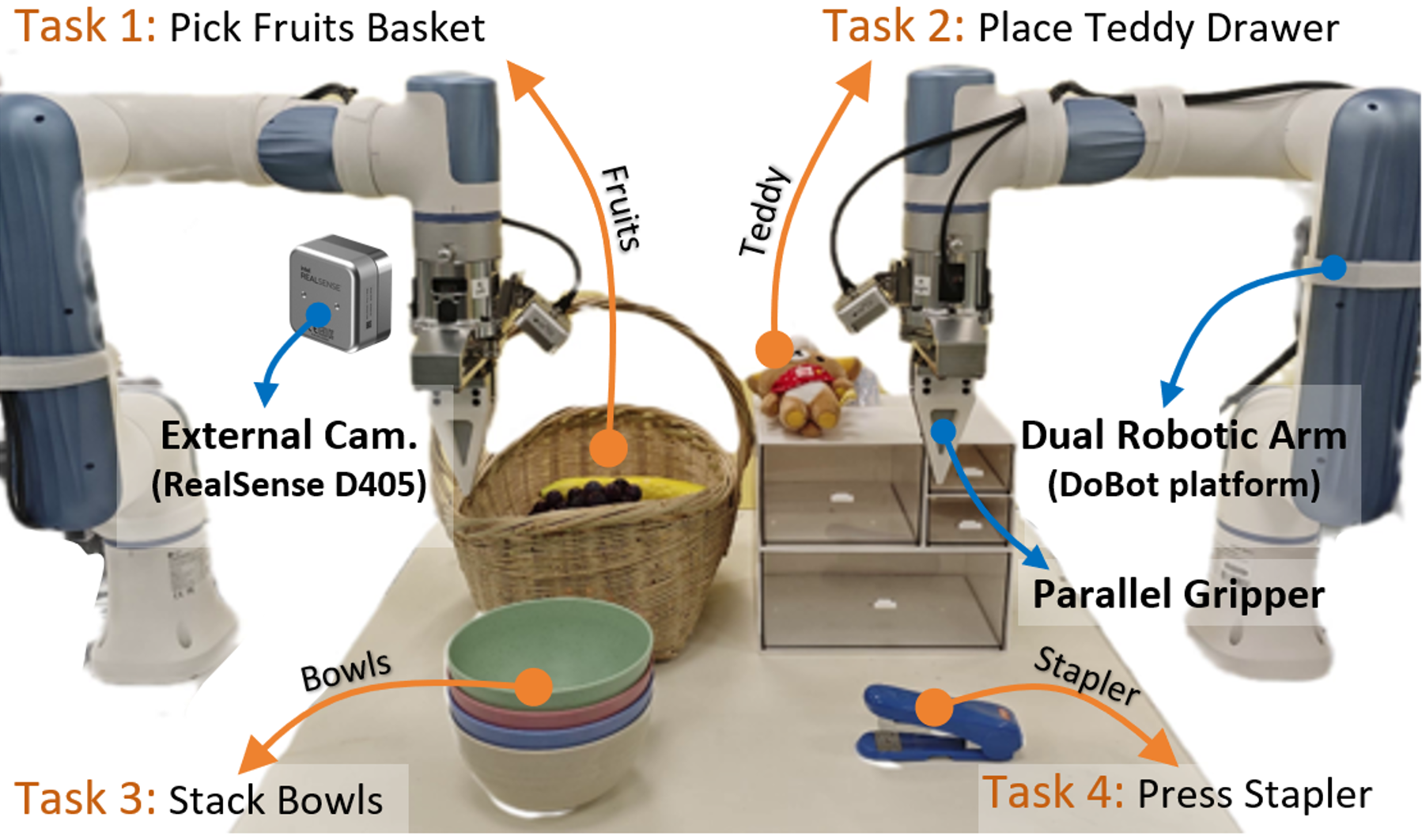}
    \caption{
        Overview of the real-world experimental setup and the four evaluation tasks.
    }
    \label{fig:real_world_setup}
\end{figure}

\subsection{Realworld Evaluation}
\label{sec:realworld_settings}
\par
As shown in Fig.~\ref{fig:real_world_setup}, we evaluate ReWeight on a dual-arm DoBot platform across four real-world tasks: \textit{Stack Bowls}; \textit{Place Teddy Drawer} (Place Teddy bear into a drawer and close it); \textit{Pick Fruit Basket} (Pick fruits and place them into a basket); and \textit{Press Stapler}. Robot data were collected via teleoperation. The sizes of the robot data and the corresponding human-data pools from EgoDex~\cite{hoque2026egodex} for retrieval are presented in Fig.~\ref{fig:task_overview}.

\par
Following the simulation setup, we compare ReWeight against two baselines: ``\textit{Robot Only}", which post-trains $\pi_{0.5}$ using only robot data, and ``\textit{Naive Mixing}", which randomly samples and incorporates human data during post-training. All models were evaluated under both clean and randomized settings. Under the clean setting, we conduct 20 trials per task, varying only the initial object positions while keeping all other environmental conditions unchanged. For \textit{Pick Fruit into Basket}, the evaluation additionally includes fruit categories present in EgoDex but absent from the robot data. Under the randomized setting, we assess robustness to visual variations using two controlled real-world perturbations: lighting variations and unseen visual distractors. For each perturbation, we conducted 10 trials per task.

\begin{figure}[t]
    \centering
    \includegraphics[width=\columnwidth]{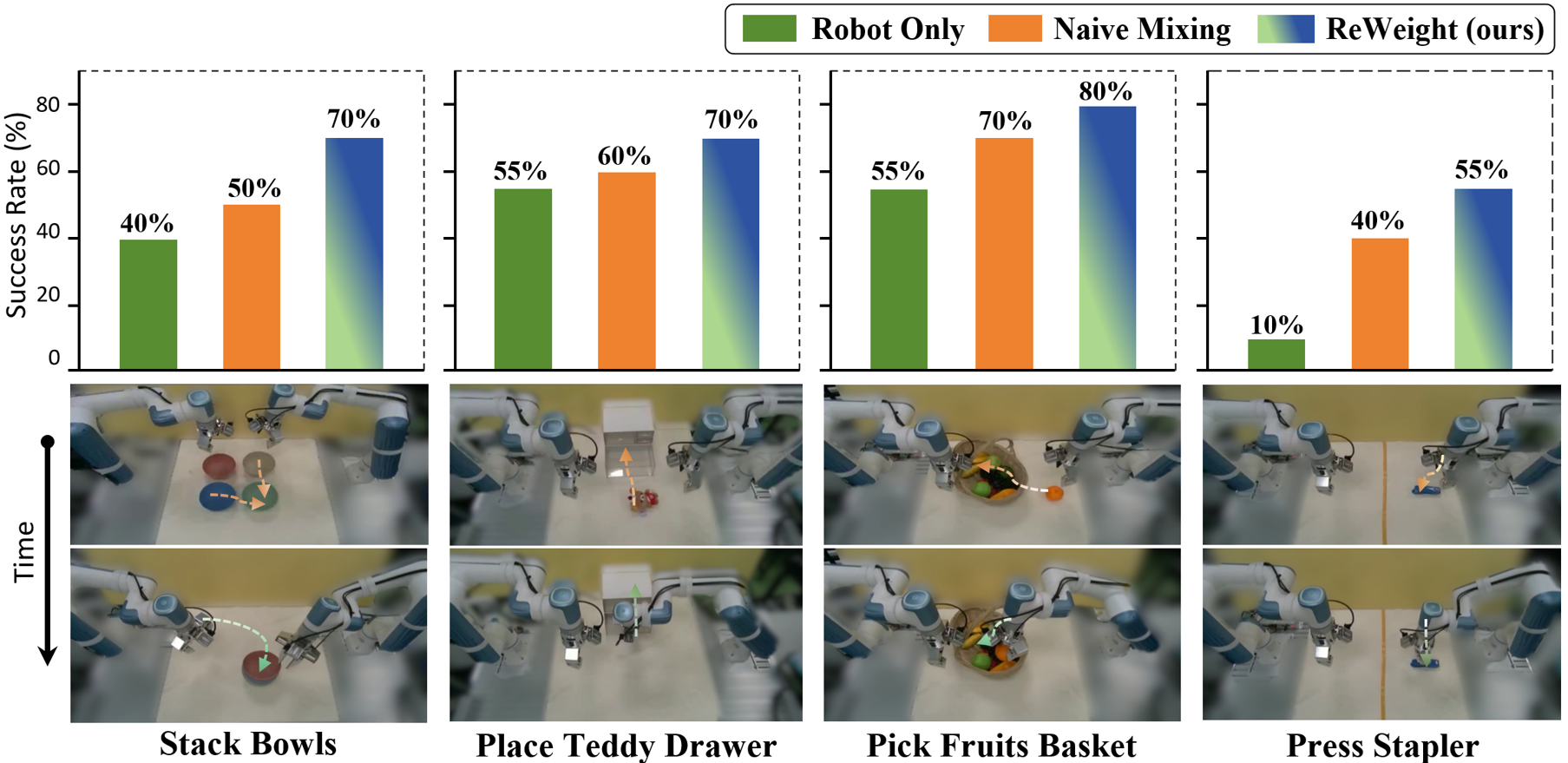}
    \caption{
    Performance comparison across four real-world tasks under the clean setting, with 20 trials conducted per task.
    }
    \label{fig:real_exp_result}
\end{figure}

\subsection{Real-world Results}
\par
As shown in Fig.~\ref{fig:real_exp_result}, ReWeight consistently outperforms both ``\textit{Robot Only}" and ``\textit{Naive Mixing}" across all four real-world tasks, improving the average success rate by 28.8 and 13.8 percentage points, respectively. This consistent improvement demonstrates that selectively retrieving and weighting egocentric human demonstrations can effectively enhance real-world robot manipulation performance. Under the randomized settings, as reported in Table~\ref{tab:real_world_generalization}, ReWeight also exhibits greater robustness to environmental perturbations. Across all four tasks, ReWeight achieves an overall success rate of 60.0\% (48/80), compared with 41.3\% (33/80) for ``\textit{Naive Mixing}" and 28.8\% (23/80) for ``\textit{Robot Only}", corresponding to improvements of 18.7 and 31.2 percentage points, respectively. The improvement is particularly pronounced on \textit{Pick Fruits Basket} task, where ReWeight succeeds in 16/20 trials, compared with 12/20 and 9/20 trials for ``\textit{Naive Mixing}" and ``\textit{Robot Only}", respectively. These results demonstrate that human-data retrieval and sample weighting improve robustness to visual variations beyond naive mixing of human and robot demonstrations. Moreover, ReWeight successfully manipulates object categories such as bananas and grapes, which are absent from the robot teleoperation data but present in the egocentric human demonstrations, indicating improved object-level generalization.

\section{Conclusion}
\label{sec:conclusion}
\par
This work introduced ReWeight, a framework for effectively incorporating egocentric human demonstrations into VLA post-training. Rather than naively mixing human and robot data, ReWeight determines both which human demonstrations to transfer and how strongly each sample should contribute to policy optimization. It learns a cross-embodiment visuomotor representation to quantify human–robot discrepancies, retrieve relevant demonstrations, and assign discrepancy-aware sample weights. In simulation, ReWeight improves the average success rate of post-trained $\pi_{0.5}$ from 39\% with robot-only data and 44\% with randomly mixed human–robot data to 57\%. In real-world experiments, it achieves 68.8\%, outperforming the two baselines by 28.8\% and 13.8\%, respectively. Under lighting changes and visual distractors, ReWeight maintains a 60.0\% success rate, compared with 28.8\% and 41.3\% for the baselines. These results demonstrate the effectiveness and robustness of ReWeight for leveraging human demonstrations in VLA post-training.

\bibliographystyle{IEEEtran}
\bibliography{IEEEabrv, reference}

\end{document}